\documentclass[letterpaper]{article} %
\usepackage{aaai2026}  %
\usepackage{times}  %
\usepackage{helvet}  %
\usepackage{courier}  %
\usepackage[hyphens]{url}  %
\usepackage{graphicx} %
\usepackage{natbib}  %
\usepackage{caption} %
\usepackage{algorithm}
\usepackage{algorithmic}
\usepackage{booktabs}
\usepackage{multirow}

\usepackage{amsmath}   
\usepackage{amssymb} 
\usepackage{bm}

\usepackage{newfloat}
\usepackage{listings}
\DeclareCaptionStyle{ruled}{labelfont=normalfont,labelsep=colon,strut=off} %
\floatstyle{ruled}
\newfloat{listing}{tb}{lst}{}
\floatname{listing}{Listing}
\nocopyright
\title{Unsupervised Real-World Super-Resolution via Rectified Flow Degradation Modelling}
\author{
Hongyang Zhou\textsuperscript{\rm 1}, Xiaobin Zhu\textsuperscript{\rm 1}, Liuling Chen\textsuperscript{\rm 1}, Junyi He\textsuperscript{\rm 2}, Jingyan Qin\textsuperscript{\rm 2}, Xu-Cheng Yin\textsuperscript{\rm 1},Xiaoxing  Zhang \textsuperscript{\rm 3}
}
\affiliations{
    
\textsuperscript{\rm 1} School of Computer and Communication Engineering, University of Science and Technology Beijing\\
\textsuperscript{\rm 2} School of Intelligence Science and Technology, University of Science and Technology Beijing\\
\textsuperscript{\rm 3}	Yizhi, China Telecom\\
        \{ilaopi,qinjingyanking\}@foxmail.com,\{zhuxiaobin,xuchengyin\}@ustb.edu.cn, \{d202310391,d202510493\}@xs.ustb.edu.cn, zhangxx7@chinatelecom.cn,  
}

\usepackage{bibentry}

\begin{document}

\maketitle

\begin{abstract}
Unsupervised real-world super-resolution (SR) faces critical challenges due to the complex, unknown degradation distributions in practical scenarios. Existing methods struggle to generalize from synthetic low-resolution (LR) and high-resolution (HR) image pairs to real-world data due to a significant domain gap. In this paper, we propose an unsupervised real-world SR method based on rectified flow to effectively capture and model real-world degradation, synthesizing LR-HR training pairs with realistic degradation. Specifically, given unpaired LR and HR images, we propose a novel Rectified Flow Degradation Module (RFDM) that introduces degradation-transformed LR (DT-LR) images as intermediaries. By modeling the degradation trajectory in a continuous and invertible manner, RFDM better captures real-world degradation and enhances the realism of generated LR images. Additionally, we propose a Fourier Prior Guided Degradation Module (FGDM) that leverages structural information embedded in Fourier phase components to ensure more precise modeling of real-world degradation. Finally, the LR images are processed by both FGDM and RFDM, producing final synthetic LR images with real-world degradation. The synthetic LR images are paired with the given HR images to train the off-the-shelf SR networks. Extensive experiments on real-world datasets demonstrate that our method significantly enhances the performance of existing SR approaches in real-world scenarios.
\end{abstract}

\section{Introduction}

Single image SR focuses on generating high-resolution (HR) images from low-resolution (LR) inputs and serves as a core task in low-level vision. While deep learning-based approaches~\cite{SRCNN, EDSR, RCAN, MambaIRv2,zhy1,zhy2,zsx1,zsx2,zsx3,en1,en2,hz1} have shown strong performance under predefined degradations like Bicubic or Gaussian, their effectiveness often declines on real-world images due to the gap between synthetic training data and complex, unknown real-world degradation~\cite{Blind_survey, Real_survey}.

\begin{figure}[t]
\centering
\includegraphics[width=1.0\columnwidth]{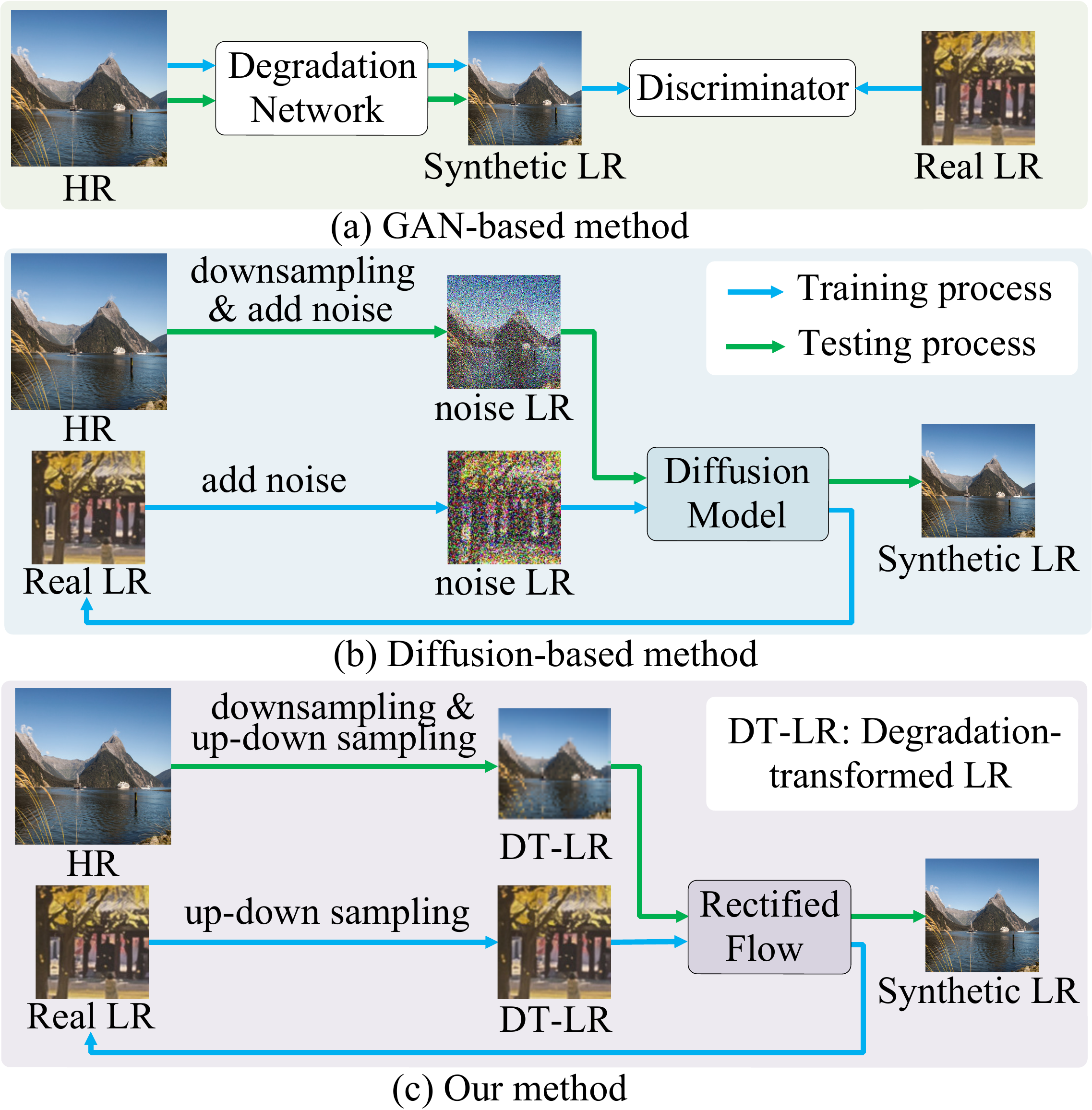} 
\caption{Illustration of different approaches for synthesizing realistic HR-LR pairs from unpaired HR-LR data.}
\label{abstract}
\end{figure}

Supervised real-world SR methods rely on either real paired datasets~\cite{RealSR, DRealSR} or synthetic degradation~\cite{BSRGAN, RealESRGAN}. While real datasets provide realistic training data, they are costly to collect and often suffer from alignment and color issues~\cite{Blind_survey, Real_survey}. Synthetic approaches are more scalable but struggle to simulate accurate degradation~\cite{Blind_survey}. Unsupervised methods address these limitations by utilizing unpaired LR-HR datasets to either train SR models directly or generate synthetic LR-HR pairs. The former~\cite{CycleGAN, UnpairedSR, DASR} often struggles with capturing the true HR distribution due to training instability~\cite{Blind_survey, SEAL}. The latter employs generative methods to synthesize LR-HR pairs, as shown in Fig.~\ref{abstract} (a) and (b), primarily using adversarial learning to capture real degradation or leveraging diffusion to learn real degradation from noisy LR images. For example, SynReal~\cite{SynReal} uses diffusion to refine noisy LR inputs from HR images, but its reliance on noise may introduce distortions in the training pairs. UDDM~\cite{UDDM} combines adversarial learning and diffusion to synthesize LR, but it is limited by the instability of adversarial training. Additionally, using diffusion to learn real degradation from extreme downsampling LR images leads to significant information loss, which impacts the quality of the synthesized LR-HR pairs. Thus, constructing realistic LR-HR pairs from unpaired data remains challenging.

Recently, flow matching~\cite{fm1} has shown great promise in image restoration~\cite{FlowIE}, especially with the development of rectified flow~\cite{rf1,fm3}. Unlike diffusion-based methods~\cite{DDPM} that rely on iterative denoising, rectified flow models a straight trajectory from a simple prior to the data distribution, leading to more efficient sampling and stable training. This makes it particularly suitable for SR tasks, as it can more effectively capture the complex degradation in real-world LR images. However, its potential in SR remains largely unexplored, especially under unpaired settings where the absence of LR-HR pairs makes degradation modeling more challenging.

In this paper, we observe that after repeated up-down sampling operations, LR images with different degradation transform into similar degradation. Based on this observation, we propose an unsupervised real-world super-resolution method using a Rectified Flow-based Degradation Module (RFDM) to effectively capture real-world degradation. Specifically,as shown in Fig.~\ref{abstract} (c), we first apply repeated up-down sampling to real-world LR images to obtain degradation-transformed LR (DT-LR) images. RFDM then learns a continuous and invertible flow transformation from DT-LR to real-world LR images, allowing it to model complex real-world degradation using only real LR data. During inference, this learned transformation is applied to DT-LR images generated from HR images, ensuring that the generated LR images follow real-world degradation. Additionally, to further enhance the accuracy of degradation modeling, we introduce a Fourier Prior Guided Degradation Module (FGDM). Leveraging the prior that degradation primarily affects the amplitude in the Fourier domain while the phase preserves structural information, FGDM refines the amplitude of DT-LR images using the phase of real LR images as structural guidance, thereby facilitating more precis degradation modeling in RFDM. Finally, the synthetic LR images generated by FGDM and RFDM, which follow real-world degradation distributions, are paired with HR images to train the SR models. Our main contributions are three-fold:

\begin{itemize}
    \item We propose an unsupervised real-world super-resolution via rectified flow degradation modelling, which effectively captures authentic degradation to synthesize realistic training data, thereby enhancing current SR methods' performance in real-world scenarios.
    \item We propose a novel Rectified Flow-based Degradation Module (RFDM) that utilizes degradation-transformed LR (DT-LR) images as intermediaries to bridge unpaired LR-HR data, effectively modeling real-world degradation via Rectified Flow’s ability to learn complex and invertible transformations.
    \item We propose a Fourier Prior Guided Degradation Module (FGDM) that leverages structural information embedded in Fourier phase components to ensure more precise modeling of real-world degradation in RFDM.
\end{itemize}

\section{Related Work}

\textbf{Single Image SR Methods}. Early SR approaches such as SRCNN~\cite{SRCNN} employ shallow convolutional networks. Subsequent methods introduce residual learning~\cite{Residual}, which enables the design of deeper architectures~\cite{VDSR,EDSR,RDN} with improved reconstruction capabilities. Attention-based models~\cite{RCAN,RBAN,HAN} are proposed to emphasize informative regions and refine feature representation. Transformer-based approaches~\cite{SwinIR,DAT} further improve performance by capturing long-range dependencies. More recently, MambaIR~\cite{MambaIRv2} leverages Vision Mamba~\cite{Mamba} to model global context efficiently with linear complexity. Despite these advances, these methods often generate perceptually smooth results lacking fine details. To address this, adversarial training~\cite{SRGAN,RealESRGAN,BSRGAN} and diffusion-based methods~\cite{Df1, Resshift} are investigated to enhance perceptual quality. Though successful on synthetic degradation, they struggle with real-world scenarios.

\noindent\textbf{Real-World SR Methods}. Supervised real-world SR methods either acquire real LR-HR pairs~\cite{RealSR, DRealSR} using specialized imaging systems, which capture authentic degradation but require costly hardware~\cite{Real_survey}, or simulate real-world degradation by enumerating the degradation operations~\cite{BSRGAN, RealESRGAN, StableSR, InvSR}, which offer scalability but suffer from inaccurate degradation simulation~\cite{Blind_survey}. Unsupervised real-world SR methods typically utilize unpaired LR-HR images to implicitly learn degradation patterns and directly train SR models~\cite{fast_gan, CycleGAN, UnpairedSR, DASR, SDFlow}. In contrast, some methods first generate paired LR-HR data from unpaired images before training. For example,  SynReal~\cite{SynReal} trains a diffusion model on real LR images to iteratively add noise to HR inputs and denoise them, producing realistic LR-HR pairs. However, this process often introduces distortions in the generated pairs. UDDM~\cite{UDDM} combines GANs and diffusion models to synthesize LR-HR pairs from extremely downsampled LR images, but the loss of fine details and the instability of adversarial learning limit their quality.

\begin{figure*}[t]
\centering
\includegraphics[width=1.0\textwidth]{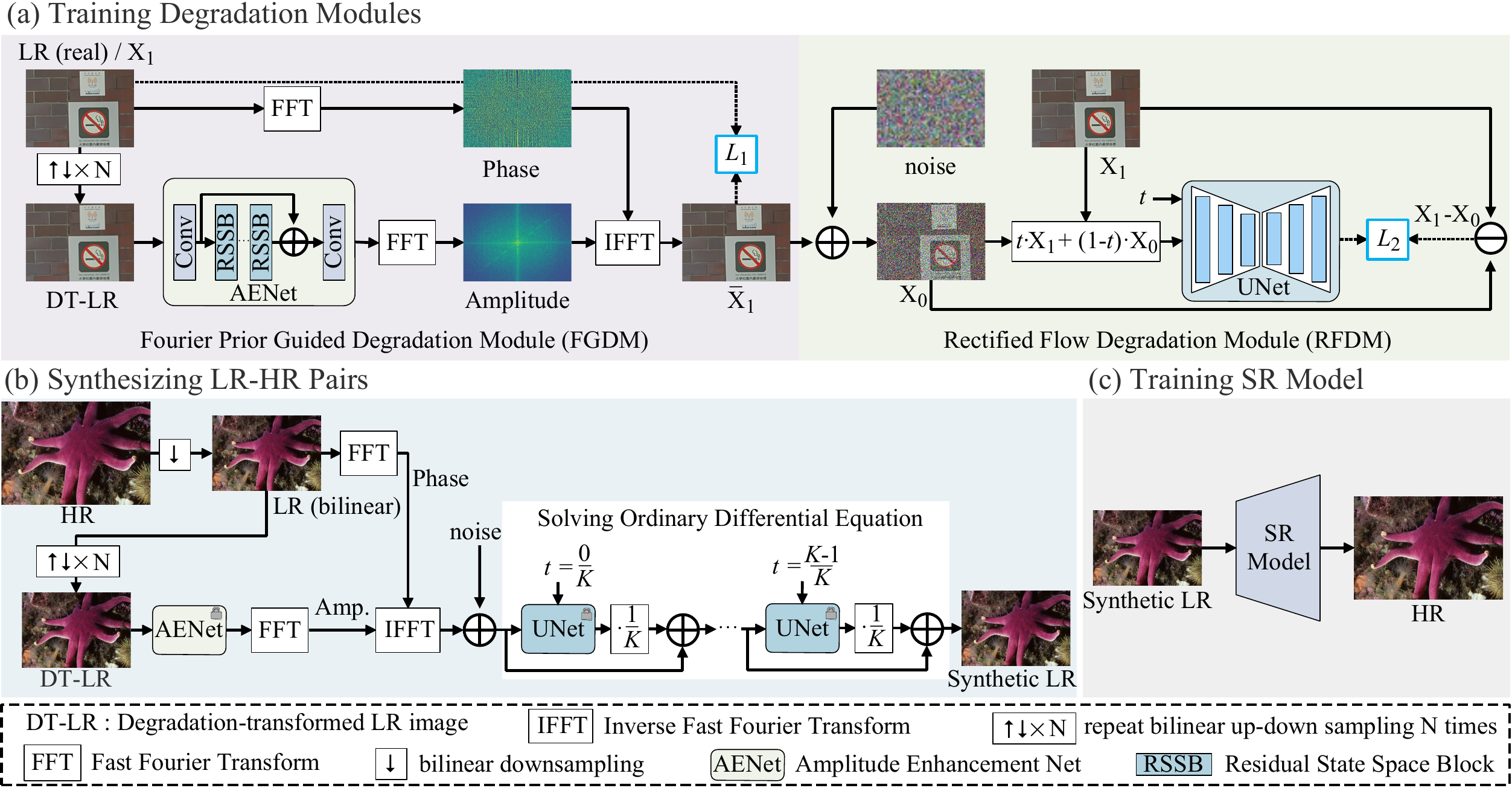} 
\caption{Overview of the proposed method. During the training phase, the Fourier Prior Degradation Module (FGDM) and Rectified Flow Degradation Module (RFDM) are trained using only real-world LR images. In the synthesis phase, the trained models generate realistic pseudo LR images. Finally, the resulting HR-LR pairs can be used to train any SR model.}
\label{arch}
\end{figure*}

\section{Our Method}

\subsection{Overview}

Our goal is to reduce the degradation gap between training and testing datasets so that SR models trained on synthetic datasets can be applied to real-world LR images. As shown in Fig.~\ref{arch}, our method mainly consists of three components: training degradation modules, synthesizing LR-HR pairs, and training the SR model.

Training degradation modules is a crucial component of our method and consists of two key parts: the Fourier Prior Guided Degradation Module (FGDM) and the Rectified Flow-based Degradation Module (RFDM). Specially, as shown in Fig.~\ref{arch} (a), we first train FGDM, which applies repeated up-down sampling to generate DT-LR images, enhances the amplitude components, and uses phase information to preserve structural details, facilitating the initial learning of realistic degradation. Building upon this, we then train RFDM, which further refines the degradation modelling by leveraging rectified flow to capture real-world degradation transformations.

Once the degradation modules (FGDM and RFDM) are optimized, as shown in Fig.~\ref{arch} (b), we synthetic LR-HR pairs by sequentially applying both modules. Specifically, the HR images are first downsampled using bilinear interpolation to obtain LR (bi) images. This LR (bi) images are then processed by the FGDM to introduce initial real-world degradations, followed by the RFDM, which further refines these degradation to more closely match real world. This pipeline ensures that the synthesized LR-HR image pairs allow the SR model trained on them to generalize effectively to real-world LR inputs.

\subsection{Rectified Flow Degradation Module}

Flow matching~\cite{fm1,fm2} formulates generative modeling as solving an ordinary differential equation (ODE):
\begin{equation}
dZ_t=v(Z_t,t)dt,
\label{eq:1}
\end{equation}
where $v$ is a time-dependent velocity field that transforms samples from a simple source distribution $P_{Z_0}$ to a target distribution $P_{Z_1}$. By integrating this ODE from $P_{Z_0}$, one can generate samples from $P_{Z_1}$. Since Eq.~\ref{eq:1} may admit multiple valid solutions, flow matching aims to learn a unique $v$ that ensures transformations between distributions. Rectified flow~\cite{fm3} defines a class of flow matching based on linear interpolation:
\begin{equation}
Z_t = tZ_1 + (1 - t)Z_0,
\label{eq:2}
\end{equation}
which yields a constant velocity vector field $dZ_t = (Z_1 - Z_0)dt$. While this provides direct linear paths between $P_{Z_0}$ and $P_{Z_1}$, it assumes access to $Z_1$ at all times $t < 1$, violating causality and limiting its applicability in generative modeling. To overcome this limitation, rectified flow adopts an alternative approach:
\begin{equation}
v(Z_t, t) = \mathbb{E}[Z_1 - Z_0 \mid Z_t],
\label{eq:causal_field}
\end{equation}
which ensures a well-posed solution to the ODE in Eq.\ref{eq:1}. Notably, solving Eq.\ref{eq:1} with $v$ often approximates the optimal transport map from  $P_{Z_0}$ to $P_{Z_1}$, particularly when applied iteratively or when the marginals are close to the optimal transport plan~\cite{fm3, fm4}. To estimate $v$, we can train $v_\theta$ with the loss criterion as:
\begin{equation}
\min_\theta\int_0^1\mathbb{E}[||(Z_1-Z_0)-v_\theta(Z_t,t)||^2]dt.
\label{eq:4}
\end{equation}

\begin{figure}[t]
\centering
\includegraphics[width=0.475\textwidth]{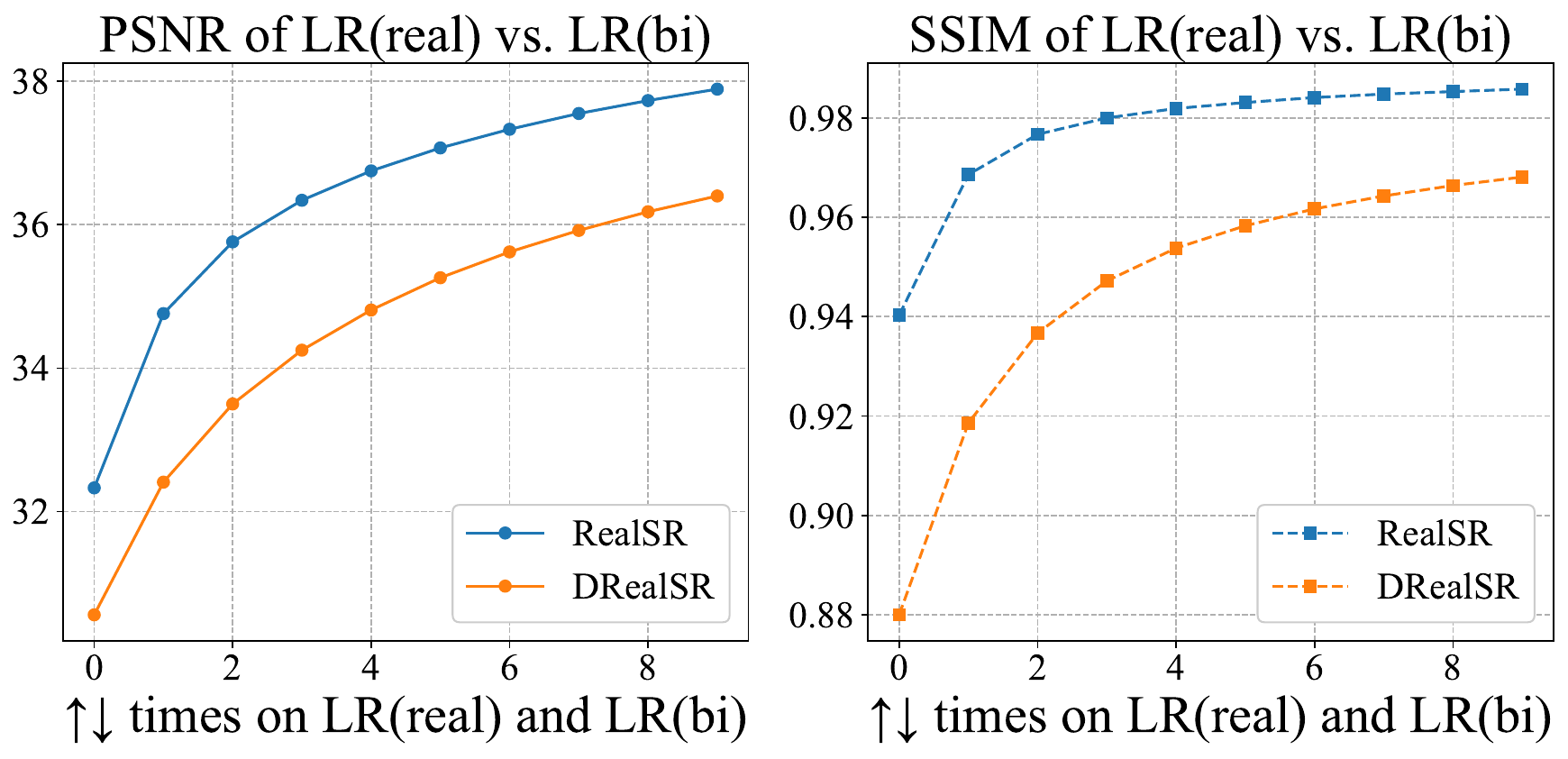}
\includegraphics[width=0.475\textwidth]{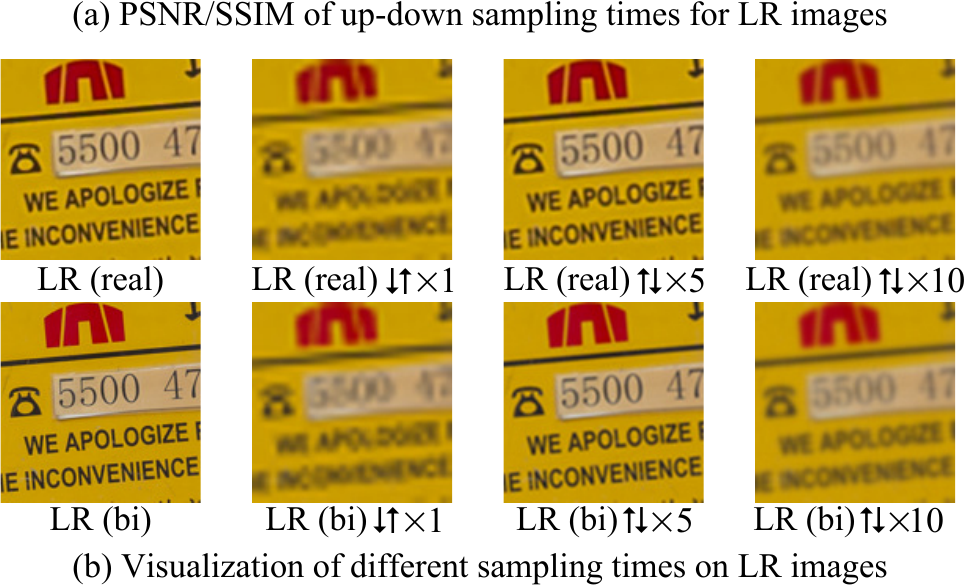} 
\caption{Illustration of degradation removal through repeated up-down sampling operation.}
\label{degradation_removal}
\end{figure}

Rectified flow provides a solid foundation for learning real-world degradation from real LR images. By mapping samples from a simple distribution to a target distribution using causal velocity fields, it effectively models real degradation. This flow process allows us to better approximate the distribution of real-world degradation in LR images. As shown in Fig.~\ref{arch} (a), after obtaining a preliminary real degraded LR $\bar X$ in stage 1, we train RFDM as following: 
\begin{gather}
\min_\theta \int_0^1 \mathbb{E}\left[\|(X_1 - X_0) - v_\theta(X_t, t)\|^2\right] dt, \label{eq:5a} \\
X_t := t X_1 + (1 - t) X_0, \label{eq:5b} \\
X_0 := \bar{X} + \lambda n, \quad n \sim \mathcal{N}(0, I). \label{eq:5c}
\end{gather}
where $\lambda$ is a hyper-parameter that controls the level of the Gaussian noise. As shown by~\cite{fm1,rf2}, adding such a noise is critical when the source and target distributions lie on low and high dimensional manifolds, respectively. Specifically, it alleviates the singularities resulting from learning a deterministic mapping between such distributions. We employ a UNet architecture to parameterize the velocity field $v_\theta$, and optimize the objective function in Eq.~\ref{eq:5a} through $L_2$ loss.

As shown in Fig.~\ref{arch} (b), to synthesize LR-HR pairs that better reflect realistic degradation, we start from initial LR images $X_0$ with preliminary real-world degradation, derived from the corresponding HR images. We then numerically solve the ODE to simulate the degradation process and obtain more realistically degraded LR images. Specifically, we solve the ODE using the Euler method with $K$ discrete steps. Starting from $X_0$, the sample is iteratively updated as follows:
\begin{equation}
X_{\frac{i+1}K} = X_{\frac iK} + \frac1K v_\theta(X_{\frac iK}, \frac iK), i=0,1,...,K-1.
\label{eq:6}
\end{equation}
After $K$ steps, we obtain the final degraded images $X_1$, which are regarded as pseudo LR images that closely approximates the real-world degradation.

\subsection{Fourier Prior Guided Degradation Module}

We observe that after repeated up-down sampling operations, LR images with different degradation transform into similar degradation. As shown in Fig.~\ref{degradation_removal} (a), we calculate the PSNR and SSIM between real-world degraded LR (real) images and bilinearly degraded LR (bi) images after different up-down sampling times. The results from two real-world datasets indicate that as the times of up-down samplings increases, both PSNR and SSIM steadily improve, suggesting that the degradation in LR images transforms towards similarity. We obtain degradation-transformed LR (DT-LR) images through repeated up-down sampling operations and use them as intermediaries to bridge unpaired LR and HR images. Besides, in Fig.\ref{degradation_removal} (b), we compare the visual result of using a single downsampling operation, as in UDDM\cite{UDDM}, where LR information is significantly lost. In contrast, even after 10 iterations of up-down sampling, our method retains considerable texture structure, which aids in preserving image details while learning real degradation.

\begin{figure}[h]
\centering
\includegraphics[width=0.475\textwidth]{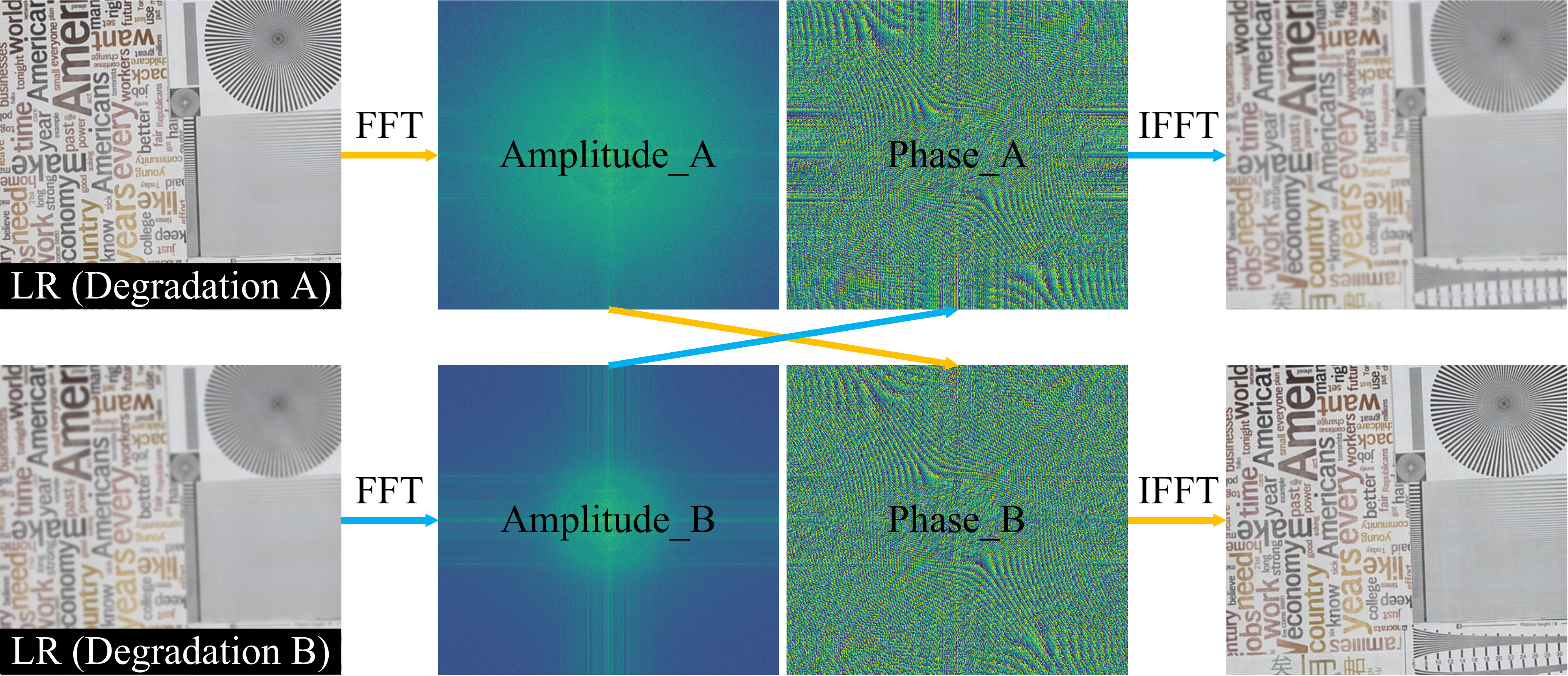} 
\caption{Illustration of amplitude and phase exchange in the Fourier domain. Swapping amplitude while keeping phase fixed shows that degradation information is mainly contained in the amplitude.}
\label{amp_pha}
\end{figure}

While repeated up-down sampling helps transform similar degradation, it inevitably leads to information loss, limiting the effectiveness of RFDM in detail recovery. To address this, we propose the Fourier Prior Guided Degradation Module (FGDM) for initial degradation modelling. As shown in Fig.\ref{amp_pha}, based on the Fourier prior\cite{fourier1,fourier2}, degradation information mainly resides in the amplitude, while structural information are preserved in the phase. As shown in Fig.~\ref{arch}, leveraging this prior, FGDM applies Fast Fourier Transform (FFT) to the DT-LR image to obtain its amplitude and phase. The amplitude is enhanced using a dedicated Amplitude Enhancement Network (AENet), while the original phase from the LR image serves as structural guidance. The enhanced amplitude and guided phase are then combined via inverse FFT (IFFT) to reconstruct a preliminary LR image with realistic degradation. AENet is composed of convolutional layers and Residual State Space Blocks (RSSB)~\cite{MambaIR} and is trained with L1 loss.

\begin{table*}[t]
\centering
\begin{tabular}{llcccccccc}
\toprule
\multirow{2}{*}{Method} & 
\multirow{2}{*}{} & 
\multicolumn{4}{c}{RealSR \cite{RealSR}} & 
\multicolumn{4}{c}{DRealSR \cite{DRealSR}} \\
\cmidrule(lr){3-6} \cmidrule(lr){7-10}
& & PSNR↑ & SSIM↑ & LPIPS↓ & FID↓ & PSNR↑ & SSIM↑ & LPIPS↓ & FID↓ \\
\midrule
SwinIR (Real-ESRGAN) & & 24.395 & 0.7760  & 0.3037 & 119.43 & 26.944 & 0.8308 & 0.3219 & 139.18 \\
SwinIR (Syn-Real) & & 25.589 & 0.7687 & 0.3835 & 163.13 & 28.301 & 0.8309 & 0.3801 & 154.59 \\
SwinIR (UDDM) & & 26.732 & 0.7913 & 0.2652 & 105.92 & 29.247 & 0.8386 & 0.2709 & 118.09 \\
SwinIR \textbf{(Ours)} & & \textbf{27.022} & \textbf{0.7981} & \textbf{0.2517} & \textbf{101.83} & \textbf{29.514} & \textbf{0.8409} & \textbf{0.2510} & \textbf{112.11} \\
\midrule
Real-ESRGAN (Real-ESRGAN) & & 25.600 & 0.7587 & 0.2749 & 138.94 & 28.549 & 0.8043 & 0.2820 & 146.94 \\
Real-ESRGAN (Syn-Real)  & & 24.341 & 0.7370 & 0.3021 & 159.44 & 27.483 & 0.7899 & 0.3306 & 171.89 \\
Real-ESRGAN (UDDM) & & 26.651 & 0.7769 & 0.2061 & 102.43 & 29.176 & 0.8032 & 0.2645 & 150.44 \\
Real-ESRGAN \textbf{(Ours)} & & \textbf{27.024} & \textbf{0.7932} & \textbf{0.1915} & \textbf{94.90} & \textbf{29.323} & \textbf{0.8051} & \textbf{0.2412} & \textbf{142.74} \\
\midrule
StableSR (Real-ESRGAN) & & 24.629 & 0.7035 & 0.3014 & 133.92 & 27.846 & 0.7412 & 0.3337 & 152.62 \\
StableSR (Syn-Real) & & 25.679 & 0.7302 & 0.3680 & 165.62 & 28.621 & 0.7952 & 0.3892 & 183.45 \\
StableSR (UDDM) & & 26.820 & 0.7768 & 0.2514 & 128.11 & 29.678 & 0.8267 & 0.2567 & 140.55 \\
StableSR \textbf{(Ours)} & & \textbf{27.128} & \textbf{0.7798} & \textbf{0.2333} & \textbf{112.45} & \textbf{29.792} & \textbf{0.8313} & \textbf{0.2396} & \textbf{132.17} \\
\bottomrule
\end{tabular}
\caption{Quantitative comparisons of the SR performance of representative models (trained with distinct data generation methods) on RealSR and DRealSR datasets. The best results are highlighted in \textbf{bold}.}
\label{tab:sr_comparison}
\end{table*}

\section{Experiments}

\subsection{Implementation Details and Datasets}

To generate DT-LR images, we apply bilinear down-up sampling 10 times. The number of RSSB blocks in AENet is set to 3, and the hyperparameter $\lambda$ in Eq.~\ref{eq:5c} is set to 0.1. We train our FGDM and RFDM using the Adam optimizer. All experiments are conducted with PyTorch 2.2.1 on NVIDIA RTX 3090 GPUs. For training, we construct unpaired HR-LR pairs. The HR dataset includes DIV2K~\cite{DIV2K}, Flickr2K~\cite{Flickr2K}, and OutdoorSceneTrain~\cite{OutdoorScence}, while the LR dataset consists of training datasets from RealSR~\cite{RealSR} and DRealSR~\cite{DRealSR}. For testing, we use the corresponding testing datasets of RealSR and DRealSR. To ensure fair comparison, we crop and evaluate the central region of each image. The resolutions of LR and HR images are 128$\times$128 and 512$\times$512, respectively. We employ PSNR, SSIM~\cite{SSIM}, LPIPS~\cite{LPIPS}, and FID~\cite{FID} to assess both the fidelity and perceptual quality of SR images.

\begin{figure}[h]
\centering
\includegraphics[width=1.0\columnwidth]{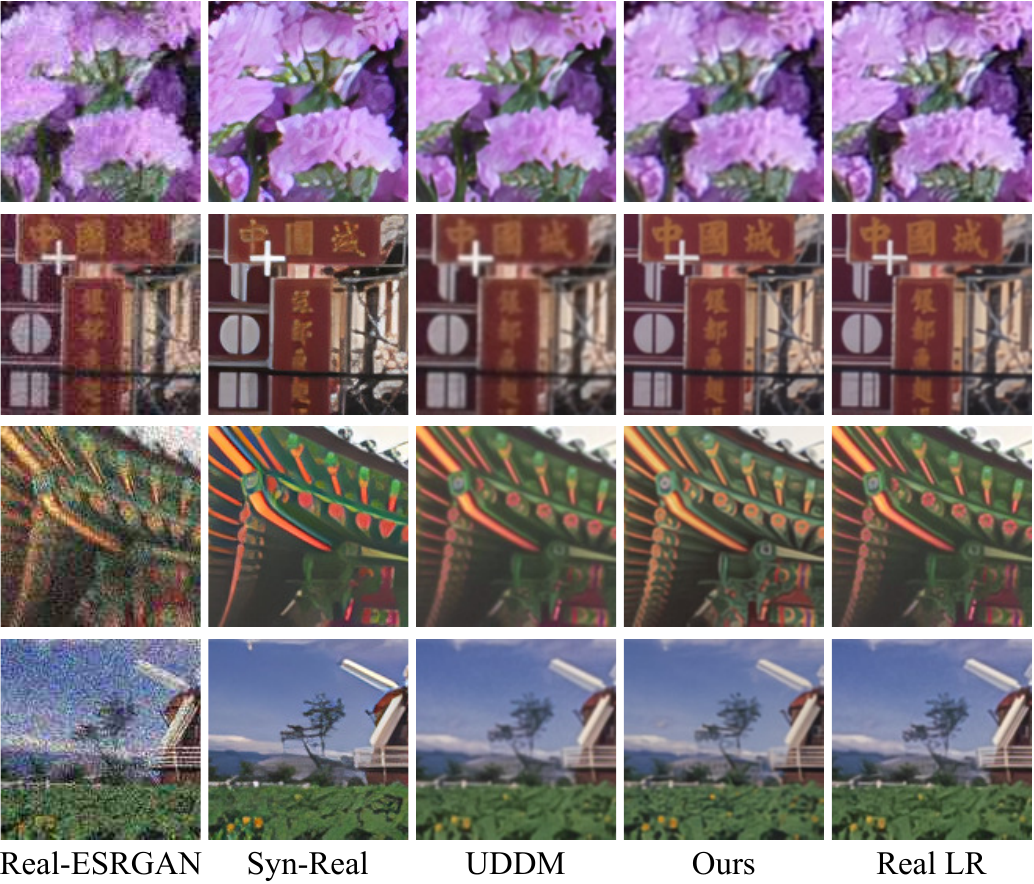} 
\caption{Qualitative comparisons for different synthetic LR images on RealSR and DRealSR datasets.}
\label{visual_LR}
\end{figure}

\begin{figure*}[t]
\centering
\includegraphics[width=2.13\columnwidth]{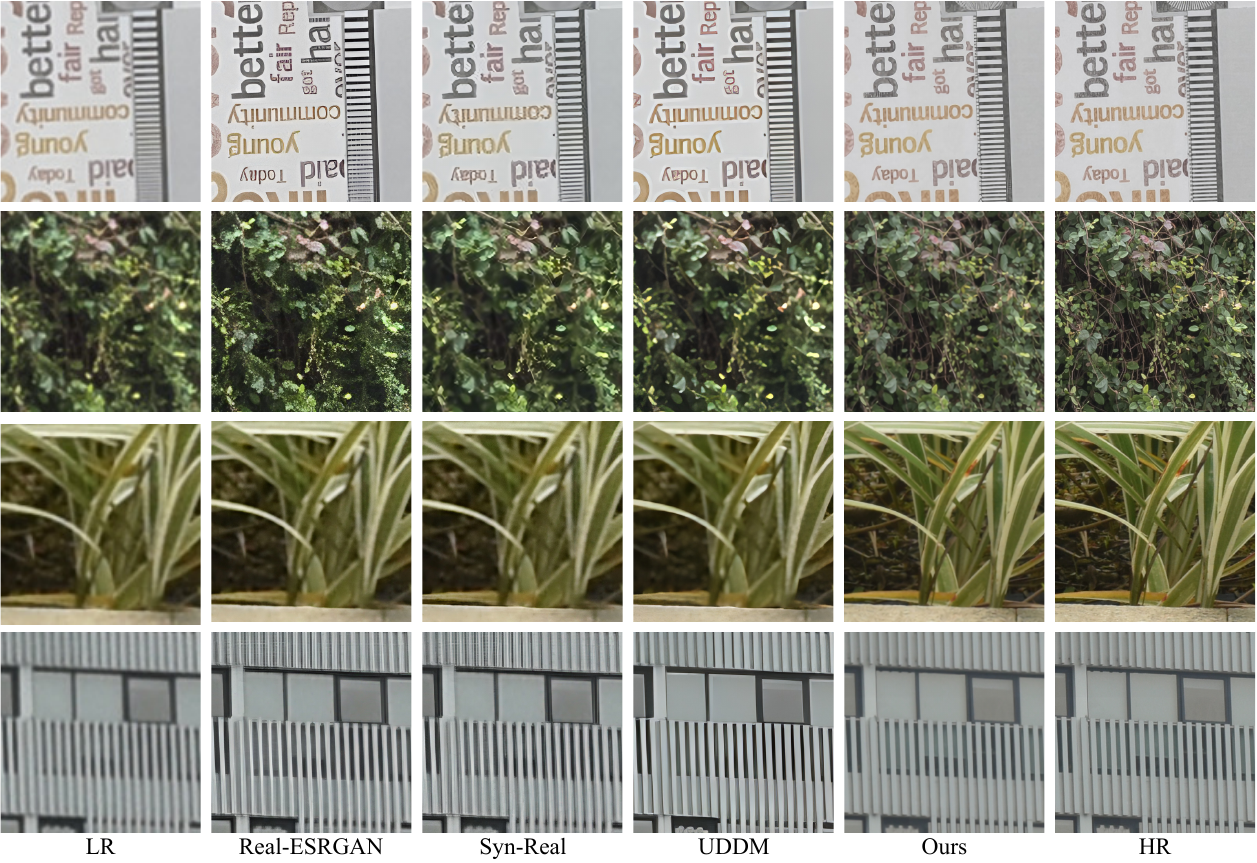} 
\caption{Qualitative results of SR models trained with different synthetic training pairs on the RealSR and DRealSR datasets.}
\label{visual}
\end{figure*}

\subsection{Quantitative Evaluation}

To evaluate the effectiveness of our method, we use the generated synthetic pairs to train several representative SR models in a supervised setting. For a fair comparison, we adopt the same selection of SR architectures as in previous work~\cite{UDDM}, including SwinIR~\cite{SwinIR}, Real-ESRGAN~\cite{RealESRGAN}, and StableSR~\cite{StableSR}, which represent transformer-based, GAN-based, and diffusion-based frameworks, respectively. Additionally, we train these SR models using training pairs generated by other methods, such as Real-ESRGAN~\cite{RealESRGAN}, Syn-Real~\cite{SynReal}, and UDDM~\cite{UDDM}, to evaluate the impact of different synthetic training pairs generation methods.

As listed in Tab.~\ref{tab:sr_comparison}, the training data synthesized by our method significantly enhances the performance of SR methods on real-world datasets. Specifically, across all four evaluation metrics, our method achieves the highest results. For example, in terms of PSNR/SSIM, our approach outperforms UDDM by 0.29 dB/0.0068 and 0.267 dB/0.0023 on the RealSR and DRealSR datasets, respectively. In comparison to StableSR, our method surpasses the Syn-Real approach by 0.1347/53.17 and 0.1496/51.28 for LPIPS/FID on RealSR and DRealSR datasets, respectively. These results validate the effectiveness of our approach in improving SR performance in real-world scenarios.

\subsection{Qualitative Evaluation}

To demonstrate the effectiveness of our method in terms of visual performance, we trained SwinIR on datasets synthesized by different methods and evaluated their performance on the RealSR and DRealSR datasets. As shown in Fig.~\ref{visual}, the first two rows display results on the RealSR dataset, while the last two rows show results on the DRealSR dataset. We observe that other methods suffer from structural distortions, especially in character and architectural line details, while our method produces fewer and less noticeable distortions. In terms of complex plant textures, the performance of other methods is less satisfactory: Real-ESRGAN introduces numerous artifacts, and Syn-Real and UDDM generate blurry results. In contrast, our method generates plant textures that are much closer to the HR images. These visual results validate the effectiveness of our approach.

To further validate our method, we performed a visual comparison of LR images synthesized by different methods from HR images. As shown in Fig.~\ref{visual_LR}, LR images synthesized by Real-ESRGAN exhibit excessive noise and artifacts, showing a significant difference from the real LR images. Syn-Real suffers from structural distortions, while UDDM shows issues with the loss of structural textures. In contrast, our method produces LR images that are much closer to the real LR images. This demonstrates the superiority of our approach in producing more realistic LR images.

\begin{table}[h]
\centering
\begin{tabular}{cccc}
\toprule
\multirow{2}{*}{FGDM} & \multirow{2}{*}{RFDM} & RealSR & DRefSR \\
 \cmidrule(lr){3-4} 
 &  & PSNR/SSIM & PSNR/SSIM \\
\midrule
 \checkmark&  & 26.395/0.7784 & 29.152/0.8263 \\
&\checkmark  & 25.221/0.7701 & 28.152/0.8215 \\
\checkmark & \checkmark & \textbf{27.022/0.7981} & \textbf{29.514/0.8409} \\
\bottomrule
\end{tabular}
\caption{Ablation study of FGDM and RFDM using the SwinIR method, with evaluation on RealSR and DRefSR.}
\label{study_modules}
\end{table}

\subsection{Ablation Study}

\noindent\textbf{Effectiveness of FGDM and RFDM.} To evaluate the effectiveness of the proposed FGDM and RFDM modules, we conduct ablation experiments, as shown in Tab.~\ref{study_modules}. When using RFDM alone, the model struggles to recover accurate results due to the lack of structural information in DT-LR images. In contrast, FGDM alone achieves better performance by leveraging Fourier priors to guide the learning of the real-world degradation. When both modules are combined, the model achieves the best results, demonstrating that FGDM and RFDM are complementary and jointly contribute to performance improvement.

\begin{figure}[h]
\centering
\includegraphics[width=1.0\columnwidth]{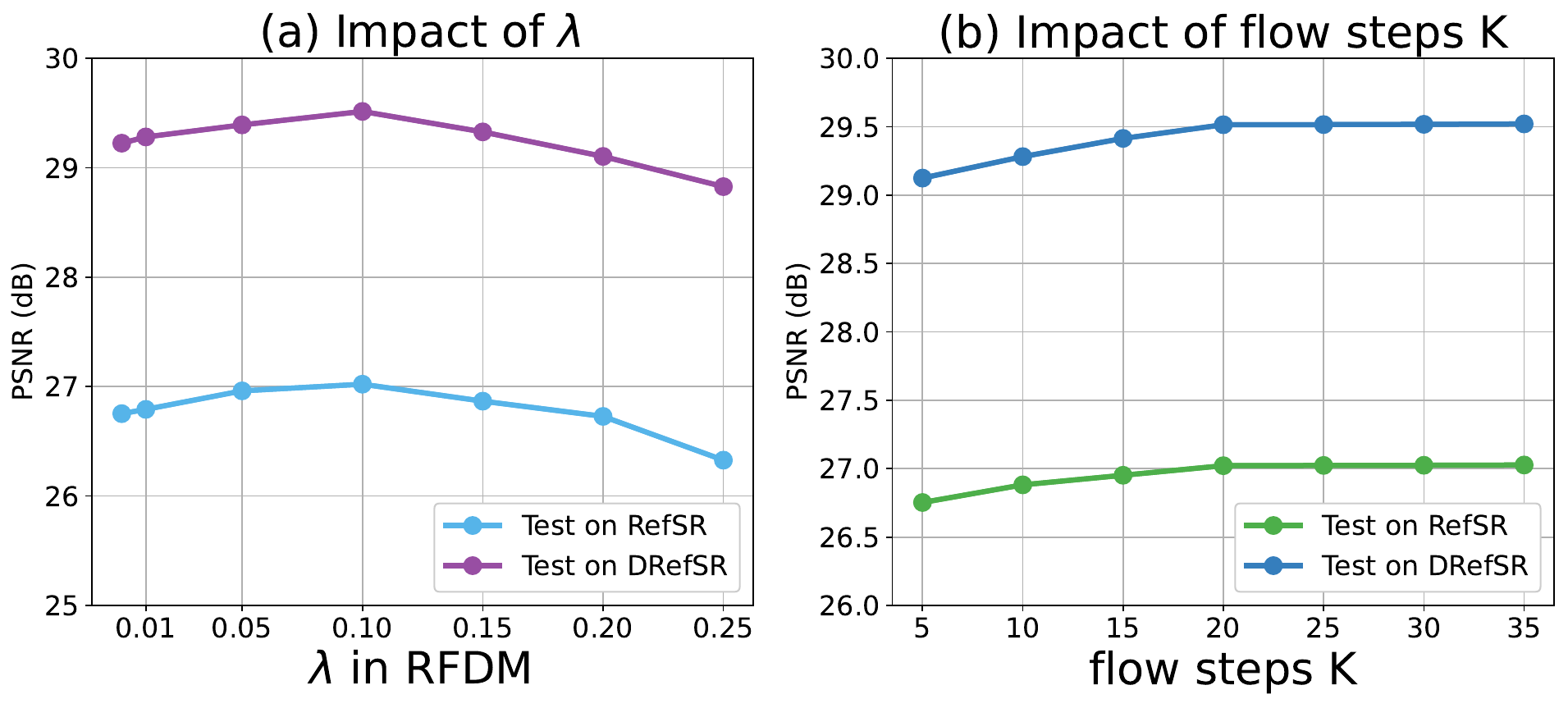} 
\caption{Ablation study of $\lambda$ and flow steps $K$ in RFDM. Testing on RealSR and DRealSR datasets using SwinIR.}
\label{study_lambda}
\end{figure}

\noindent\textbf{About $\lambda$ in RFDM.} We investigate the impact of the noise injection ratio $\lambda$ by varying its value. As shown in Fig.~\ref{study_lambda}(a), when $\lambda=0$, no noise is added, and thus the rectified flow has limited effect, resulting in minimal performance gain. As $\lambda$ increases, performance improves, indicating that moderate noise facilitates better flow refinement. However, when $\lambda$ exceeds 0.1, excessive noise can exacerbate image distortion, leading to a decline in PSNR. Therefore, we choose $\lambda=0.1$ as a trade-off between effectiveness and stability.

\noindent\textbf{About flow steps $K$ in RFDM}. To evaluate the impact of flow steps $K$ in the ODE during inference, we conduct experiments using SwinIR. As shown in Fig.~\ref{study_lambda}(b), PSNR increases with larger $K$ on both testing datasets until $K=20$, beyond which the improvement becomes marginal. Therefore, we set $K=20$ to balance performance and computational efficiency.

\begin{figure}[h]
\centering
\includegraphics[width=1.0\columnwidth]{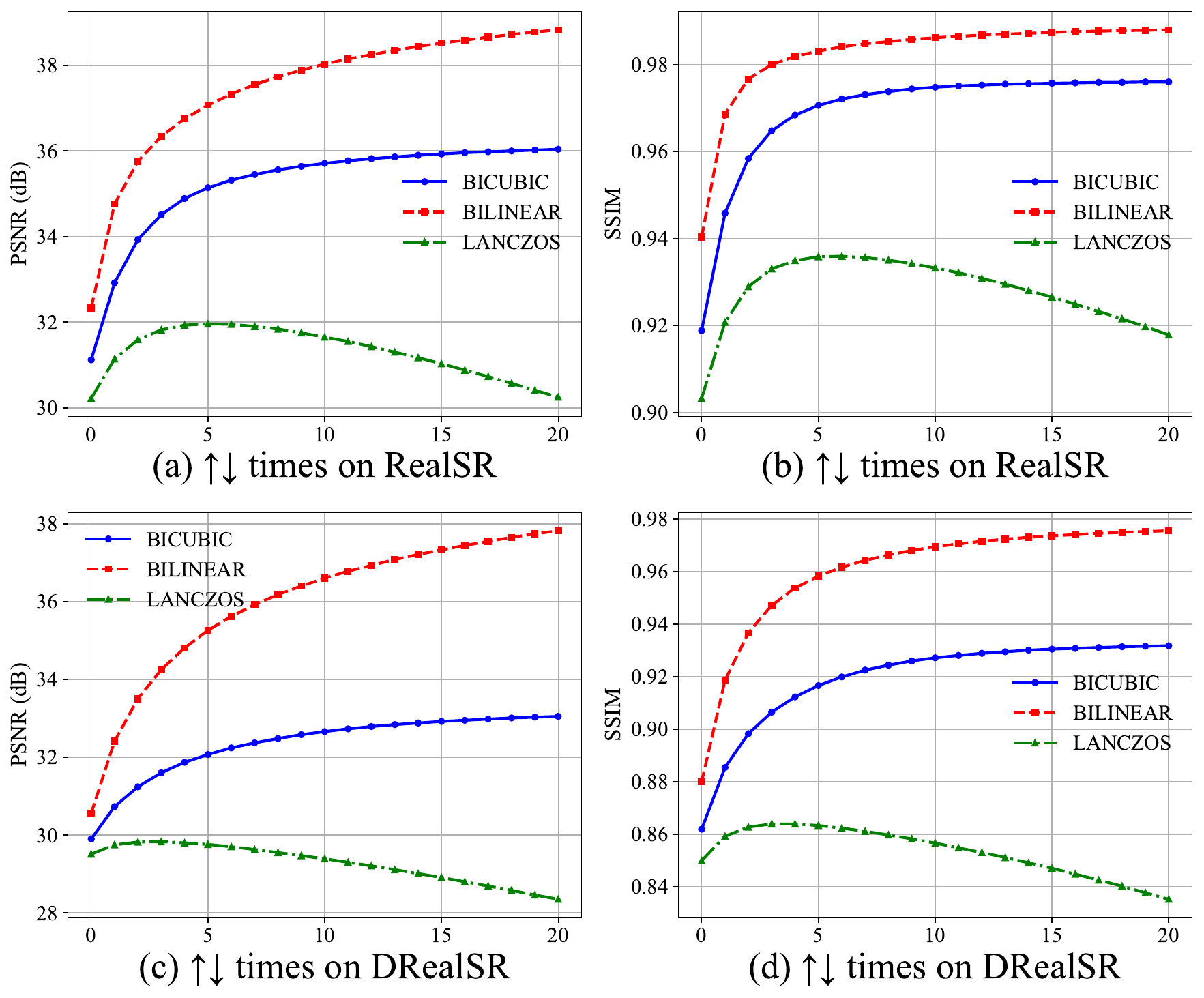} 
\caption{About different sampling methods, we first generate LR images and then perform different numbers of up-down sampling operations on both generated LR images and real LR images. Finally, we calculate the PSNR/SSIM.}
\label{study_sample}
\end{figure}

\noindent\textbf{About DT-LR images}. To investigate the impact of different sampling strategies, we use three common methods: Bicubic, Bilinear, and Lanczos. First, we select a sampling strategy to generate LR images, then apply the different number of up-down sampling operations on both the generated LR images and the real LR images. Finally, we calculate the PSNR/SSIM between the two sets of images. As shown in Fig.~\ref{study_sample}, the results on the RealSR and DRealSR datasets reveal that with Lanczos, PSNR and SSIM initially improve with an increasing number of up-down sampling operations, but quickly decline. In contrast, Bicubic shows slow improvement in PSNR and SSIM, without reaching the desired peak. This suggests that Lanczos and Bicubic cannot effectively transform LR images with different degradations into similar degradations. Bilinear, however, best meets our requirements. After only a few up-down sampling operations, the degradation of LR images gradually becomes consistent. Although PSNR and SSIM improve with more operations, LR information loss becomes significant (More analysis in \textbf{Supp. Materials}). We selected 10 operations, balancing texture preservation and bridging unpaired LR and HR images.

\begin{table}[h]
\centering
\begin{tabular}{ccc}
\toprule
\multirow{2}{*}{method} & RealSR & DRefSR \\
 \cmidrule(lr){2-3} 
   & PSNR/SSIM & PSNR/SSIM \\
\midrule
 RGDM (w/o FP) & 26.712/0.7891& 29.211/0.8335 \\
 RGDM (with FP) & \textbf{27.022/0.7981} & \textbf{29.514/0.8409} \\
\bottomrule
\end{tabular}
\caption{Ablation study of Fourier prior (FP) in FGDM using SwinIR method, with evaluation on RealSR and DRefSR.}
\label{study_fourier}
\end{table}

\noindent\textbf{About Fourier prior in FGDM}.To verify the effectiveness of the Fourier prior, we conducted ablation experiments as listed in Tab.~\ref{study_fourier}. The results demonstrate that incorporating the Fourier prior (FP) in FGDM leads to significant improvements in both RealSR and DRefSR datasets. Specifically, the PSNR and SSIM values for RGDM with FP are higher than those without FP, achieving 27.022/0.7981 and 29.514/0.8409 on RealSR and DRefSR, respectively. These improvements highlight the importance of the Fourier prior in enhancing the performance of the FGDM.

\section{Conclusion}
In this paper, we proposed an unsupervised real-world super-resolution via rectified flow degradation modelling, synthesizing LR-HR training pairs with realistic degradation. Our approach introduces two key modules: the Rectified Flow Degradation Module (RFDM) and the Fourier Prior Guided Degradation Module (FGDM). RFDM captures real-world degradation by modelling the degradation trajectory in a continuous and invertible manner, using degradation-transformed LR (DT-LR) images as intermediaries to bridge unpaired LR-HR pairs. Meanwhile, FGDM leverages the structural information embedded in Fourier phase components to ensure a more precise modelling of degradation. By utilizing both modules, we generate synthetic LR images that closely resemble real-world degradations, which are then paired with HR images for training the off-the-shelf SR network. Extensive experiments on real-world datasets show that our method offers a promising solution for enhancing SR performance in real-world applications.

\bibliography{aaai2026}

\begin{thebibliography}{56}
\providecommand{\natexlab}[1]{#1}

\bibitem[{Agustsson and Timofte(2017)}]{DIV2K}
Agustsson, E.; and Timofte, R. 2017.
\newblock Ntire 2017 challenge on single image super-resolution: Dataset and
  study.
\newblock In \emph{Proceedings of the IEEE Conference on Computer Vision and
  Pattern Recognition Workshops}, 126--135.

\bibitem[{Albergo and Vanden-Eijnden(2023)}]{fm1}
Albergo, M.~S.; and Vanden-Eijnden, E. 2023.
\newblock Building normalizing flows with stochastic interpolants.
\newblock In \emph{International Conference on Learning Representations}.

\bibitem[{Bulat, Yang, and Tzimiropoulos(2018)}]{fast_gan}
Bulat, A.; Yang, J.; and Tzimiropoulos, G. 2018.
\newblock To learn image super-resolution, use a gan to learn how to do image
  degradation first.
\newblock In \emph{Proceedings of the European Conference on Computer Vision},
  185--200.

\bibitem[{Cai et~al.(2019)Cai, Zeng, Yong, Cao, and Zhang}]{RealSR}
Cai, J.; Zeng, H.; Yong, H.; Cao, Z.; and Zhang, L. 2019.
\newblock Toward real-world single image super-resolution: A new benchmark and
  a new model.
\newblock In \emph{Proceedings of the IEEE/CVF International Conference on
  Computer Vision}, 3086--3095.

\bibitem[{Chen et~al.(2022)Chen, He, Qing, Wu, Ren, Sheriff, and
  Zhu}]{Real_survey}
Chen, H.; He, X.; Qing, L.; Wu, Y.; Ren, C.; Sheriff, R.~E.; and Zhu, C. 2022.
\newblock Real-world single image super-resolution: A brief review.
\newblock \emph{Information Fusion}, 79: 124--145.

\bibitem[{Chen et~al.(2025)Chen, Yao, Li, Pei, Zhao, and Ren}]{UDDM}
Chen, Y.; Yao, M.; Li, W.; Pei, R.; Zhao, J.; and Ren, W. 2025.
\newblock Unsupervised Diffusion-Based Degradation Modeling for Real-World
  Super-Resolution.
\newblock In \emph{Proceedings of the AAAI Conference on Artificial
  Intelligence}, volume~39, 2348--2356.

\bibitem[{Chen et~al.(2023)Chen, Zhang, Gu, Kong, Yang, and Yu}]{DAT}
Chen, Z.; Zhang, Y.; Gu, J.; Kong, L.; Yang, X.; and Yu, F. 2023.
\newblock Dual Aggregation Transformer for Image Super-Resolution.
\newblock In \emph{International Conference on Computer Vision}, 12312--12321.

\bibitem[{Dai et~al.(2019)Dai, Zha, Jiang, and Xia}]{RBAN}
Dai, T.; Zha, H.; Jiang, Y.; and Xia, S.-T. 2019.
\newblock Image super-resolution via residual block attention networks.
\newblock In \emph{IEEE/CVF International Conference on Computer Vision
  Workshops}.

\bibitem[{Dong et~al.(2014)Dong, Loy, He, and Tang}]{SRCNN}
Dong, C.; Loy, C.~C.; He, K.; and Tang, X. 2014.
\newblock Learning a deep convolutional network for image super-resolution.
\newblock In \emph{European Conference on Computer Vision}, 184--199.

\bibitem[{Gu and Dao(2024)}]{Mamba}
Gu, A.; and Dao, T. 2024.
\newblock Mamba: Linear-Time Sequence Modeling with Selective State Spaces.
\newblock In \emph{Conference on Language Modeling}.

\bibitem[{Guo et~al.(2025)Guo, Guo, Zha, Zhang, Li, Dai, Xia, and
  Li}]{MambaIRv2}
Guo, H.; Guo, Y.; Zha, Y.; Zhang, Y.; Li, W.; Dai, T.; Xia, S.-T.; and Li, Y.
  2025.
\newblock Mambairv2: Attentive state space restoration.
\newblock In \emph{Proceedings of the Computer Vision and Pattern Recognition
  Conference}, 28124--28133.

\bibitem[{Guo et~al.(2024)Guo, Li, Dai, Ouyang, Ren, and Xia}]{MambaIR}
Guo, H.; Li, J.; Dai, T.; Ouyang, Z.; Ren, X.; and Xia, S.-T. 2024.
\newblock Mambair: A simple baseline for image restoration with state-space
  model.
\newblock In \emph{European Conference on Computer Vision}, 222--241.

\bibitem[{Han et~al.(2024)Han, Zhu, Yang, Zhou, Qin, and Yin}]{hz1}
Han, Z.; Zhu, X.; Yang, C.; Zhou, H.; Qin, J.; and Yin, X.-C. 2024.
\newblock Exploring Stable Meta-Optimization Patterns via Differentiable
  Reinforcement Learning for Few-Shot Classification.
\newblock In \emph{Proceedings of the 32nd ACM International Conference on
  Multimedia}, 1701--1710.

\bibitem[{He et~al.(2016)He, Zhang, Ren, and Sun}]{Residual}
He, K.; Zhang, X.; Ren, S.; and Sun, J. 2016.
\newblock Deep residual learning for image recognition.
\newblock In \emph{IEEE Conference on Computer Vision and Pattern Recognition},
  770--778.

\bibitem[{Heusel et~al.(2017)Heusel, Ramsauer, Unterthiner, Nessler, and
  Hochreiter}]{FID}
Heusel, M.; Ramsauer, H.; Unterthiner, T.; Nessler, B.; and Hochreiter, S.
  2017.
\newblock Gans trained by a two time-scale update rule converge to a local nash
  equilibrium.
\newblock \emph{Advances in Neural Information Processing Systems}, 30.

\bibitem[{Ho, Jain, and Abbeel(2020)}]{DDPM}
Ho, J.; Jain, A.; and Abbeel, P. 2020.
\newblock Denoising diffusion probabilistic models.
\newblock \emph{Advances in Neural Information Processing Systems}, 33:
  6840--6851.

\bibitem[{Kim, Kwon~Lee, and Mu~Lee(2016)}]{VDSR}
Kim, J.; Kwon~Lee, J.; and Mu~Lee, K. 2016.
\newblock Accurate image super-resolution using very deep convolutional
  networks.
\newblock In \emph{Proceedings of the IEEE Conference on Computer Vision and
  Pattern Recognition}, 1646--1654.

\bibitem[{Ledig et~al.(2017)Ledig, Theis, Husz{\'a}r, Caballero, Cunningham,
  Acosta, Aitken, Tejani, Totz, Wang et~al.}]{SRGAN}
Ledig, C.; Theis, L.; Husz{\'a}r, F.; Caballero, J.; Cunningham, A.; Acosta,
  A.; Aitken, A.; Tejani, A.; Totz, J.; Wang, Z.; et~al. 2017.
\newblock Photo-realistic single image super-resolution using a generative
  adversarial network.
\newblock In \emph{Proceedings of the IEEE Conference on Computer Vision and
  Pattern Recognition}, 4681--4690.

\bibitem[{Liang et~al.(2021)Liang, Cao, Sun, Zhang, Van~Gool, and
  Timofte}]{SwinIR}
Liang, J.; Cao, J.; Sun, G.; Zhang, K.; Van~Gool, L.; and Timofte, R. 2021.
\newblock Swinir: Image restoration using swin transformer.
\newblock In \emph{Proceedings of the IEEE/CVF International Conference on
  Computer Vision Workshops}, 1833--1844.

\bibitem[{Lim et~al.(2017)Lim, Son, Kim, Nah, and Mu~Lee}]{EDSR}
Lim, B.; Son, S.; Kim, H.; Nah, S.; and Mu~Lee, K. 2017.
\newblock Enhanced deep residual networks for single image super-resolution.
\newblock In \emph{Proceedings of the IEEE Conference on Computer Vision and
  Pattern Recognition Workshops}, 136--144.

\bibitem[{Lipman et~al.(2023)Lipman, Chen, Ben-Hamu, Nickel, and Le}]{fm2}
Lipman, Y.; Chen, R.~T.; Ben-Hamu, H.; Nickel, M.; and Le, M. 2023.
\newblock Flow Matching for Generative Modeling.
\newblock In \emph{International Conference on Learning Representations}.

\bibitem[{Liu et~al.(2022)Liu, Liu, Gu, Qiao, and Dong}]{Blind_survey}
Liu, A.; Liu, Y.; Gu, J.; Qiao, Y.; and Dong, C. 2022.
\newblock Blind image super-resolution: A survey and beyond.
\newblock \emph{IEEE Transactions on Pattern Analysis and Machine
  Intelligence}, 45(5): 5461--5480.

\bibitem[{Liu(2022)}]{rf1}
Liu, Q. 2022.
\newblock Rectified flow: A marginal preserving approach to optimal transport.
\newblock \emph{arXiv preprint arXiv:2209.14577}.

\bibitem[{Liu, Gong, and Liu(2023)}]{fm3}
Liu, X.; Gong, C.; and Liu, Q. 2023.
\newblock Flow Straight and Fast: Learning to Generate and Transfer Data with
  Rectified Flow.
\newblock In \emph{International Conference on Learning Representations}.

\bibitem[{Maeda(2020)}]{UnpairedSR}
Maeda, S. 2020.
\newblock Unpaired image super-resolution using pseudo-supervision.
\newblock In \emph{Proceedings of the IEEE/CVF conference on computer vision
  and pattern recognition}, 291--300.

\bibitem[{Morimitsu et~al.(2025)Morimitsu, Zhu, Cesar, Ji, and Yin}]{en2}
Morimitsu, H.; Zhu, X.; Cesar, R.~M.; Ji, X.; and Yin, X.-C. 2025.
\newblock DPFlow: Adaptive Optical Flow Estimation with a Dual-Pyramid
  Framework.
\newblock In \emph{Proceedings of the Computer Vision and Pattern Recognition
  Conference}, 17810--17820.

\bibitem[{Morimitsu et~al.(2024)Morimitsu, Zhu, Ji, and Yin}]{en1}
Morimitsu, H.; Zhu, X.; Ji, X.; and Yin, X.-C. 2024.
\newblock Recurrent partial kernel network for efficient optical flow
  estimation.
\newblock In \emph{Proceedings of the AAAI Conference on Artificial
  Intelligence}, volume~38, 4278--4286.

\bibitem[{Nehete et~al.(2024)Nehete, Monga, Kaushik, and Kaushik}]{fourier1}
Nehete, H.; Monga, A.; Kaushik, P.; and Kaushik, B.~K. 2024.
\newblock Fourier Prior-Based Two-Stage Architecture for Image Restoration.
\newblock In \emph{Proceedings of the IEEE/CVF Conference on Computer Vision
  and Pattern Recognition Workshop}, 6014--6023.

\bibitem[{Niu et~al.(2020)Niu, Wen, Ren, Zhang, Yang, Wang, Zhang, Cao, and
  Shen}]{HAN}
Niu, B.; Wen, W.; Ren, W.; Zhang, X.; Yang, L.; Wang, S.; Zhang, K.; Cao, X.;
  and Shen, H. 2020.
\newblock Single image super-resolution via a holistic attention network.
\newblock In \emph{European Conference on Computer Vision}, 191--207.

\bibitem[{Ohayon, Michaeli, and Elad(2025)}]{rf2}
Ohayon, G.; Michaeli, T.; and Elad, M. 2025.
\newblock Posterior-Mean Rectified Flow: Towards Minimum MSE Photo-Realistic
  Image Restoration.
\newblock In \emph{International Conference on Learning Representations}.

\bibitem[{Rombach et~al.(2022)Rombach, Blattmann, Lorenz, Esser, and
  Ommer}]{Df1}
Rombach, R.; Blattmann, A.; Lorenz, D.; Esser, P.; and Ommer, B. 2022.
\newblock High-resolution image synthesis with latent diffusion models.
\newblock In \emph{Proceedings of the IEEE/CVF conference on computer vision
  and pattern recognition}, 10684--10695.

\bibitem[{Sun and Chen(2024)}]{SDFlow}
Sun, W.; and Chen, Z. 2024.
\newblock Learning many-to-many mapping for unpaired real-world image
  super-resolution and downscaling.
\newblock \emph{IEEE Transactions on Pattern Analysis and Machine
  Intelligence}.

\bibitem[{Timofte et~al.(2017)Timofte, Agustsson, Van~Gool, Yang, and
  Zhang}]{Flickr2K}
Timofte, R.; Agustsson, E.; Van~Gool, L.; Yang, M.-H.; and Zhang, L. 2017.
\newblock Ntire 2017 challenge on single image super-resolution: Methods and
  results.
\newblock In \emph{Proceedings of the IEEE Conference on Computer Vision and
  Pattern Recognition Workshops}, 114--125.

\bibitem[{Tong et~al.(2024)Tong, Fatras, Malkin, Huguet, Zhang, Rector-Brooks,
  Wolf, and Bengio}]{fm4}
Tong, A.; Fatras, K.; Malkin, N.; Huguet, G.; Zhang, Y.; Rector-Brooks, J.;
  Wolf, G.; and Bengio, Y. 2024.
\newblock Improving and generalizing flow-based generative models with
  minibatch optimal transport.
\newblock \emph{Transactions on Machine Learning Research}, 1--34.

\bibitem[{Wang et~al.(2024)Wang, Yue, Zhou, Chan, and Loy}]{StableSR}
Wang, J.; Yue, Z.; Zhou, S.; Chan, K.~C.; and Loy, C.~C. 2024.
\newblock Exploiting diffusion prior for real-world image super-resolution.
\newblock \emph{International Journal of Computer Vision}, 132(12): 5929--5949.

\bibitem[{Wang et~al.(2021)Wang, Xie, Dong, and Shan}]{RealESRGAN}
Wang, X.; Xie, L.; Dong, C.; and Shan, Y. 2021.
\newblock Real-esrgan: Training real-world blind super-resolution with pure
  synthetic data.
\newblock In \emph{Proceedings of the IEEE/CVF International Conference on
  Computer Vision}, 1905--1914.

\bibitem[{Wang et~al.(2018)Wang, Yu, Dong, and Loy}]{OutdoorScence}
Wang, X.; Yu, K.; Dong, C.; and Loy, C.~C. 2018.
\newblock Recovering realistic texture in image super-resolution by deep
  spatial feature transform.
\newblock In \emph{Proceedings of the IEEE Conference on Computer Vision and
  Pattern Recognition}, 606--615.

\bibitem[{Wang et~al.(2004)Wang, Bovik, Sheikh, and Simoncelli}]{SSIM}
Wang, Z.; Bovik, A.~C.; Sheikh, H.~R.; and Simoncelli, E.~P. 2004.
\newblock Image quality assessment: from error visibility to structural
  similarity.
\newblock \emph{IEEE Transactions on Image Processing}, 13(4): 600--612.

\bibitem[{Wei et~al.(2020)Wei, Xie, Lu, Zhan, Ye, Zuo, and Lin}]{DRealSR}
Wei, P.; Xie, Z.; Lu, H.; Zhan, Z.; Ye, Q.; Zuo, W.; and Lin, L. 2020.
\newblock Component divide-and-conquer for real-world image super-resolution.
\newblock In \emph{European Conference on Computer Vision}, 101--117.

\bibitem[{Wei et~al.(2021)Wei, Gu, Li, Timofte, Jin, and Song}]{DASR}
Wei, Y.; Gu, S.; Li, Y.; Timofte, R.; Jin, L.; and Song, H. 2021.
\newblock Unsupervised real-world image super resolution via domain-distance
  aware training.
\newblock In \emph{Proceedings of the IEEE/CVF Conference on Computer Vision
  and Pattern Recognition}, 13385--13394.

\bibitem[{Yang et~al.(2023)Yang, Ren, Zhang et~al.}]{SynReal}
Yang, T.; Ren, P.; Zhang, L.; et~al. 2023.
\newblock Synthesizing realistic image restoration training pairs: A diffusion
  approach.
\newblock \emph{arXiv preprint arXiv:2303.06994}.

\bibitem[{Yuan et~al.(2018)Yuan, Liu, Zhang, Zhang, Dong, and Lin}]{CycleGAN}
Yuan, Y.; Liu, S.; Zhang, J.; Zhang, Y.; Dong, C.; and Lin, L. 2018.
\newblock Unsupervised image super-resolution using cycle-in-cycle generative
  adversarial networks.
\newblock In \emph{Proceedings of the IEEE Conference on Computer Vision and
  Pattern Recognition Workshops}, 701--710.

\bibitem[{Yue, Liao, and Loy(2025)}]{InvSR}
Yue, Z.; Liao, K.; and Loy, C.~C. 2025.
\newblock Arbitrary-steps image super-resolution via diffusion inversion.
\newblock In \emph{Proceedings of the Computer Vision and Pattern Recognition
  Conference}, 23153--23163.

\bibitem[{Yue, Wang, and Loy(2023)}]{Resshift}
Yue, Z.; Wang, J.; and Loy, C.~C. 2023.
\newblock Resshift: Efficient diffusion model for image super-resolution by
  residual shifting.
\newblock \emph{Advances in Neural Information Processing Systems}, 36:
  13294--13307.

\bibitem[{Zhang et~al.(2021)Zhang, Liang, Van~Gool, and Timofte}]{BSRGAN}
Zhang, K.; Liang, J.; Van~Gool, L.; and Timofte, R. 2021.
\newblock Designing a practical degradation model for deep blind image
  super-resolution.
\newblock In \emph{Proceedings of the IEEE/CVF international conference on
  computer vision}, 4791--4800.

\bibitem[{Zhang et~al.(2018{\natexlab{a}})Zhang, Isola, Efros, Shechtman, and
  Wang}]{LPIPS}
Zhang, R.; Isola, P.; Efros, A.~A.; Shechtman, E.; and Wang, O.
  2018{\natexlab{a}}.
\newblock The unreasonable effectiveness of deep features as a perceptual
  metric.
\newblock In \emph{Proceedings of the IEEE conference on computer vision and
  pattern recognition}, 586--595.

\bibitem[{Zhang et~al.(2024)Zhang, Yang, Zhu, Zhou, Wang, and Yin}]{zsx3}
Zhang, S.-X.; Yang, C.; Zhu, X.; Zhou, H.; Wang, H.; and Yin, X.-C. 2024.
\newblock Inverse-like antagonistic scene text spotting via reading-order
  estimation and dynamic sampling.
\newblock \emph{IEEE Transactions on Image Processing}, 33: 825--839.

\bibitem[{Zhang et~al.(2022{\natexlab{a}})Zhang, Zhu, Chen, Hou, and
  Yin}]{zsx1}
Zhang, S.-X.; Zhu, X.; Chen, L.; Hou, J.-B.; and Yin, X.-C. 2022{\natexlab{a}}.
\newblock Arbitrary shape text detection via segmentation with probability
  maps.
\newblock \emph{IEEE transactions on pattern analysis and machine
  intelligence}, 45(3): 2736--2750.

\bibitem[{Zhang et~al.(2022{\natexlab{b}})Zhang, Zhu, Hou, Yang, and
  Yin}]{zsx2}
Zhang, S.-X.; Zhu, X.; Hou, J.-B.; Yang, C.; and Yin, X.-C. 2022{\natexlab{b}}.
\newblock Kernel proposal network for arbitrary shape text detection.
\newblock \emph{IEEE transactions on neural networks and learning systems},
  34(11): 8731--8742.

\bibitem[{Zhang et~al.(2023)Zhang, Li, Chen, Zhang, Qiao, Wu, and Dong}]{SEAL}
Zhang, W.; Li, X.; Chen, X.; Zhang, X.; Qiao, Y.; Wu, X.-M.; and Dong, C. 2023.
\newblock SEAL: A Framework for Systematic Evaluation of Real-World
  Super-Resolution.
\newblock In \emph{International Conference on Learning Representations}.

\bibitem[{Zhang et~al.(2018{\natexlab{b}})Zhang, Li, Li, Wang, Zhong, and
  Fu}]{RCAN}
Zhang, Y.; Li, K.; Li, K.; Wang, L.; Zhong, B.; and Fu, Y. 2018{\natexlab{b}}.
\newblock Image super-resolution using very deep residual channel attention
  networks.
\newblock In \emph{Proceedings of the European Conference on Computer Vision},
  286--301.

\bibitem[{Zhang et~al.(2018{\natexlab{c}})Zhang, Tian, Kong, Zhong, and
  Fu}]{RDN}
Zhang, Y.; Tian, Y.; Kong, Y.; Zhong, B.; and Fu, Y. 2018{\natexlab{c}}.
\newblock Residual dense network for image super-resolution.
\newblock In \emph{Proceedings of the IEEE Conference on Computer Vision and
  Pattern Recognition}, 2472--2481.

\bibitem[{Zhao et~al.(2024)Zhao, Cai, Dong, and Hu}]{fourier2}
Zhao, C.; Cai, W.; Dong, C.; and Hu, C. 2024.
\newblock Wavelet-based fourier information interaction with frequency
  diffusion adjustment for underwater image restoration.
\newblock In \emph{Proceedings of the IEEE/CVF Conference on Computer Vision
  and Pattern Recognition}, 8281--8291.

\bibitem[{Zhou et~al.(2025)Zhou, Zhu, Qin, Xu, Cesar-Jr, and Yin}]{zhy2}
Zhou, H.; Zhu, X.; Qin, J.; Xu, Y.; Cesar-Jr, R.~M.; and Yin, X.-C. 2025.
\newblock Multi-Scale Texture Fusion for Reference-Based Image
  Super-Resolution: New Dataset and Solution.
\newblock \emph{International Journal of Computer Vision}, 1--22.

\bibitem[{Zhou et~al.(2023)Zhou, Zhu, Zhu, Han, Zhang, Qin, and Yin}]{zhy1}
Zhou, H.; Zhu, X.; Zhu, J.; Han, Z.; Zhang, S.-X.; Qin, J.; and Yin, X.-C.
  2023.
\newblock Learning correction filter via degradation-adaptive regression for
  blind single image super-resolution.
\newblock In \emph{Proceedings of the IEEE/CVF International Conference on
  Computer Vision}, 12365--12375.

\bibitem[{Zhu et~al.(2024)Zhu, Zhao, Li, Tang, Zhou, and Lu}]{FlowIE}
Zhu, Y.; Zhao, W.; Li, A.; Tang, Y.; Zhou, J.; and Lu, J. 2024.
\newblock Flowie: Efficient image enhancement via rectified flow.
\newblock In \emph{Proceedings of the IEEE/CVF Conference on Computer Vision
  and Pattern Recognition}, 13--22.

\end{thebibliography}

\end{document}